# An Uncertainty-Aware Dynamic Decision Framework for Progressive Multi-Omics Integration in Classification Tasks

Nan Mu[a,b,c,#], Hongbo Yang[a,#], and Chen Zhao[d,*]

[a]*College of Computer Science, Sichuan Normal University, Chengdu, Sichuan 610101, China.*
[b]*Visual Computing and Virtual Reality Key Laboratory of Sichuan, Sichuan Normal University, Chengdu, Sichuan 610068, China.*
[c]*Education Big Data Collaborative Innovation Center of Sichuan 2011, Chengdu, Sichuan 610101, China.*
[d]*Department of Computer Science, College of Computing and Software Engineering, Kennesaw State University, Marietta, GA 30060, USA.*

## Abstract

**Background and Objective:** High-throughput multi-omics technologies have proven invaluable for elucidating disease mechanisms and enabling early diagnosis by integrating multi-layered molecular data to reveal complex disease networks and heterogeneity. However, two key challenges hinder their practical implementation. First, the complexity of disease mechanisms involves coordinated molecular interactions that single-omics approaches often fail to capture comprehensively. Second, the high cost of multi-omics profiling imposes a significant economic burden, with over-reliance on full-omics data potentially leading to unnecessary resource consumption. To address these issues, we propose an uncertainty-aware, multi-view dynamic decision framework for omics data classification that aims to achieve high diagnostic accuracy while minimizing testing costs.

**Methodology:** At the single-omics level, we refine the activation functions of neural networks to generate Dirichlet distribution parameters, utilizing subjective logic to quantify both the belief masses (probabilities) and uncertainty mass of classification results. Belief mass reflects the support of a specific omics modality for a disease class, while the uncertainty parameter captures limitations in data quality and model discriminability, providing a more trustworthy basis for decision-making. At the multi-omics level, we employ a fusion strategy based on Dempster-Shafer theory to integrate heterogeneous modalities, leveraging their complementarity to boost diagnostic accuracy and robustness. A dynamic decision mechanism is then applied: omics data are incrementally introduced for each patient until either all data sources are utilized or the model's confidence exceeds a predefined threshold—potentially before all data sources are utilized. This enables precise, patient-specific predictions while reducing unnecessary tests.

---

[#]N. Mu and H. Yang contributed equally to this work.

[*]Corresponding author to provide e-mail: czhao4@kennesaw.edu (C. Zhao).

**Results and Conclusion:** We evaluate our approach on four benchmark multi-omics datasets—ROSMAP, LGG, BRCA, and KIPAN. In three datasets, over 50% of cases achieved accurate classification using a single omics modality, effectively reducing redundant testing. Meanwhile, our method maintains diagnostic performance comparable to full-omics models and preserves essential biological insights. This framework introduces a novel "on-demand testing" paradigm into precision medicine, enabling intelligent resource allocation and reducing healthcare costs, which is especially valuable in resource-limited clinical environments.

## 1. Introduction

High-throughput omics technologies, encompassing genomics, transcriptomics, proteomics, and metabolomics (Hasin, Seldin, & Lusis, 2017), enable rapid and large-scale profiling of biological molecules. These technologies have revolutionized biomedical research by facilitating multi-layered analysis of disease onset and progression, offering unprecedented insights into complex molecular mechanisms (Dar, et al., 2023). Moreover, the ability to detect early-stage biomarkers through high-throughput omics profiling has opened new avenues for timely diagnosis and personalized therapeutic interventions (Orsini, Diquigiovanni, & Bonora, 2023; Song, et al., 2025).

Despite these advancements, the application of multi-omics data in clinical and translational settings faces two critical challenges. First, the acquisition of multi-omics data remains prohibitively expensive (Mullin, 2022; Zhang, 2024). The costs associated with instrumentation, reagents, and computational infrastructure, combined with the need for highly skilled personnel and stringent experimental protocols, impose significant barriers to widespread adoption (Vitorino, 2024). Second, from a data analysis perspective, existing integration methods—particularly those based on neural networks and machine learning—exhibit inherent limitations. While neural networks excel at capturing nonlinear dependencies and extracting high-dimensional features, their sensitivity to data imperfections such as noise, missing values, and class imbalance can undermine predictive reliability (Picard, Scott-Boyer, Bodein, Périn, & Droit, 2021). Machine learning models, though generally more robust to overfitting, often struggle with adaptability and stability in complex and heterogeneous data environments (Fouché & Zinovyev, 2023).

To address these challenges, we propose a dynamic decision-making framework that adaptively determines the number and combination of omics modalities required for each prediction task. This approach aims to maintain high diagnostic performance while minimizing unnecessary data acquisition, thereby reducing both cost and patient burden (Han, Zhang, Fu, & Zhou, 2022). Furthermore, we introduce an evidence-based uncertainty estimation mechanism to evaluate prediction reliability across multiple omics views. By incorporating uncertainty quantification into the classification pipeline, the model is better equipped to handle data imperfections and ensure robust decision-making even under suboptimal conditions (Flores, et al., 2023; Zhang, 2024), e.g., noise, missing values, and class imbalance.

The primary contributions of this work are as follows:

- An uncertainty-aware, multi-view dynamic decision framework for omics classification is proposed. It balances diagnostic accuracy with cost-effectiveness by quantifying uncertainty at the single-omics level and dynamically integrating additional modalities only when necessary.
- A novel evidential fusion strategy is developed to estimate predictive uncertainty by aggregating multi-view information at the belief level. Unlike traditional feature- or output-level fusion, this approach allows independent evaluation of each modality while capturing joint uncertainty across views.
- A staged dynamic decision mechanism is employed to progressively incorporate omics modalities (e.g., single-, dual-, or full-view), guided by confidence thresholds. This enables efficient, patient-specific testing strategies tailored to the informativeness of available data.
- Comprehensive experimental validation is conducted on four public multi-omics datasets (ROSMAP, LGG, BRCA, and KIPAN), demonstrating that the proposed method achieves superior classification performance, robustness, and reliability compared to state-of-the-art approaches.

## 2. Related Work

*2.1.* Uncertainty-Based Learning

Uncertainty estimation plays a critical role in evaluating model confidence and guiding reliable decision-making (Kendall & Gal, 2017). Bayesian Neural Networks (BNNs), originally introduced by MacKay (MacKay, 1992) and Neal (Neal, 2012), model network parameters as probability distributions to capture epistemic uncertainty. Despite their theoretical appeal, BNNs often rely on Markov Chain Monte Carlo (MCMC) methods for inference, which impose high computational costs and hinder their scalability in large-scale multi-omics settings (Gal & Ghahramani, 2016; Mobiny, et al., 2021). To address this, Gal and Ghahramani (Gal & Ghahramani, 2016) proposed Monte Carlo Dropout (MC-Dropout), which approximates model uncertainty by applying stochastic dropout during both training and inference, using multiple forward passes to estimate predictive variance. While MC-Dropout is more efficient than MCMC-based BNNs, its effectiveness is sensitive to the choice of dropout rate and the number of samples used; improper settings may result in unreliable confidence estimates, especially when generalizing to unseen data (Hasan, Hossain, Rahman, & Nahavandi, 2023).

In contrast, our study adopts an evidential deep learning framework grounded in subjective logic (Jøsang, 2016). This approach reformulates classification as a Dirichlet distribution parameter estimation task (Sensoy, Kaplan, & Kandemir, 2018). The model outputs non-negative evidence values in a single forward pass, which are then mapped to the concentration parameters of a Dirichlet distribution (Chen, Gao, & Xu, 2024). This formulation allows simultaneous quantification of epistemic uncertainty (model ignorance) and aleatoric uncertainty (data noise) (Deng, Chen, Yu, Liu, & Heng, 2023). Crucially, this evidential method eliminates the need for multiple stochastic passes, thereby enhancing computational efficiency while improving the reliability of uncertainty estimation.

*2.2.* Dynamic Decision-Making

Dynamic decision-making plays a vital role in reducing diagnostic costs and improving efficiency. However, most existing multi-omics integration approaches—such as MOGONET (Wang, et al., 2021) and CLCLSA (Zhao, et al., 2024)—require complete data across all omics modalities prior to prediction. While effective in accuracy, this "full-omics" strategy is inefficient and costly in clinical settings. Many patients, particularly low-risk cases, may be accurately diagnosed using only a subset of modalities. Nevertheless, conventional models enforce comprehensive testing, leading to unnecessary financial burdens and overutilization of resources (Ma, et al., 2024).

To address this limitation, we propose a confidence-driven dynamic decision framework based on progressive data integration. Instead of collecting all omics modalities upfront, our method selectively incorporates additional data only when model confidence is insufficient (Wu & Xie, 2024). Specifically, we utilize the Dirichlet strength parameter $s_v$ as a measure of uncertainty, and define adaptive thresholds $t_1$ and $t_2$ to guide a stepwise diagnostic workflow: *single-view → dual-view → full-view* (Espinel-Ríos, López, & Avalos, 2025). This staged strategy ensures that further data collection is triggered only when necessary, enabling efficient use of resources while preserving diagnostic reliability.

## 3. Methodology

We consider a scenario involving $N$ subjects and $M$ types of omics data. Each subject is represented by a unique set of features $\mathcal{X}_i = \{x_i^{(1)}, \cdots, x_i^{(j)}, \cdots, x_i^{(N)}\}$, $i \in \{1, \cdots, M\}$, where $x_i^{(j)}$ denotes the feature vector of the $j$-th subject under the $i$-th omics modality. Initially, we utilize only one omics modality $x_i^{(j)}$ as input. If the prediction confidence is sufficiently high at this stage, the rest modalities are no longer required. Otherwise, we incrementally incorporate additional omics modalities until either high-confidence predictions are achieved or all modalities have been utilized, thereby enhancing predictive trustworthiness.

### 3.1. Quantifying Uncertainty with Evidential Deep Learning

This subsection introduces an evidential deep learning framework designed to quantify classification uncertainty from multiple perspectives. As shown in Figure 1, the proposed model not only estimates class probabilities but also captures the overall predictive uncertainty associated with each decision. Our approach is rooted in subjective logic, which offers a principled probabilistic framework for modeling and reasoning under uncertainty.

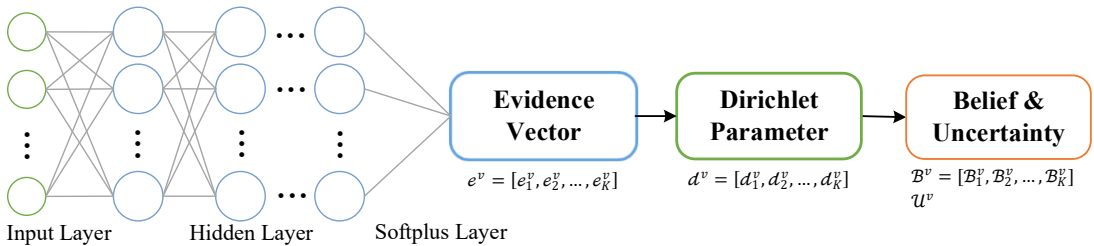

**Figure 1.** Workflow of evidential deep learning for estimating belief masses $\mathcal{B}^v$ and uncertainty mass $\mathcal{U}^v$ from single-modal omics data: From multilayer perceptron (MLP) outputs to subjective opinions.

Subjective logic (Jøsang, 2016) offers a mathematically rigorous framework for integrating Dirichlet distribution parameters with belief functions, allowing for the joint modeling of class-specific support and overall predictive uncertainty. In our approach, we leverage the Dirichlet distribution (Cover, 1999) to represent a probabilistic model over $K$ classes, where each parameter reflects the strength of evidence in favor of a particular class $k$. These parameters collectively govern the shape of the predictive distribution and encode the model's uncertainty. When one class dominates in evidence, the model exhibits strong confidence in that prediction; conversely, when all evidence values are low or similar, the model expresses uncertainty due to insufficient discriminative support.

To realize this mechanism, we employ a Multilayer Perceptron (MLP) to output a non-negative evidence vector. Rather than using the standard Softmax activation, which produces normalized probabilities but fails to represent uncertainty, we apply the Softplus function, defined as $\text{Softplus}(x) = \ln(1 + e^x)$, to ensure strictly positive outputs. This activation generates the evidence vector $e^v = [e_1^v, e_2^v, \ldots, e_K^v]$, where $e_k^v$ denotes the evidence supporting class class $k$ under view $v$. The corresponding Dirichlet parameter $d_k^v \in [d_1^v, d_2^v, \ldots, d_K^v]$ are then constructed as:

$$d_k^v = e_k^v + 1. \quad (1)$$

This construction guarantees that all Dirichlet parameters are strictly positive, establishing a well-formed prior for uncertainty modeling and avoiding the instability caused by negative outputs. Notably, the use of Softplus not only improves numerical robustness but also enhances the model's capacity to express epistemic uncertainty, making it well-suited for evidence-based reasoning in multi-omics classification.

Following subjective logic (Jøsang, 2016), we compute the belief masses $\mathcal{B}_k^v \in [\mathcal{B}_1^v, \mathcal{B}_2^v, \ldots, \mathcal{B}_K^v]$ and the uncertainty mass $\mathcal{U}^v$, subject to the constraint:

$$\mathcal{U}^v + \sum_{k=1}^{K} \mathcal{B}_k^v = 1, \quad (2)$$

where both $\mathcal{U}^v > 0$ and $\mathcal{B}_k^v > 0$. The belief masses and uncertainty mass are computed as:

$$\mathcal{B}_k^v = \frac{e_k^v}{S^v} = \frac{d_k^v - 1}{S^v}, \quad (3)$$

$$\mathcal{U}^v = \frac{K}{S^v}. \quad (4)$$

The term $S^v$, referred to as the Dirichlet strength, quantifies the total amount of evidence.

$$S^v = \sum_{k=1}^{K} (d_k^v) = \sum_{k=1}^{K} (e_k^v + 1). \quad (5)$$

A higher evidence value for a specific class increases the corresponding belief mass, while lower total evidence results in a higher uncertainty mass—indicating reduced confidence in the model's prediction.

By analyzing the distribution of belief masses, the model can make uncertainty-aware decisions. When the uncertainty mass $\mathcal{U}^v$ is high, it suggests that the model is unsure about its prediction under the current perspective. In such cases, additional perspectives or data modalities can be incorporated to reduce ambiguity and enhance the reliability of the final decision.

*3.2. Multi-View Fusion with Dempster–Shafer Theory*

This subsection introduces the application of Dempster-Shafer Theory (DST) (Dempster, 2008) for fusing multi-view information under uncertainty. The corresponding model architecture is illustrated in Figure 2. In a single-view setting, each perspective is independently evaluated to extract its associated evidence and uncertainty for a given prediction. When the uncertainty from a particular view exceeds a predefined threshold, it indicates that the evidence from that view alone is insufficient to support a confident decision. In such cases, it becomes necessary to integrate complementary information from other views to achieve a more reliable and comprehensive prediction.

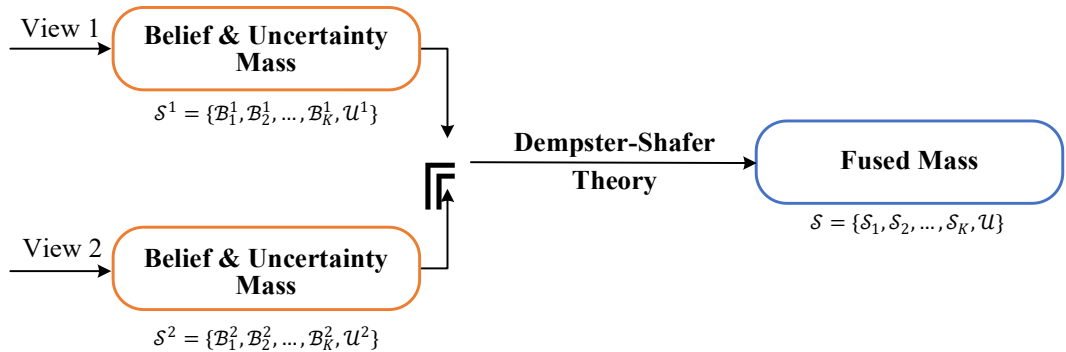

**Figure 2.** Multi-view opinion fusion using dempster-shafer theory.

To address this, we adopt DST as a principled framework for multi-view opinion integration. DST offers a powerful mathematical framework for evidential reasoning under uncertainty, especially when dealing with incomplete, imprecise, or ambiguous data. Unlike conventional probabilistic models that require full information, DST allows reasoning based on partial evidence and provides mechanisms for combining belief from multiple sources. This makes DST particularly suitable for high-stakes applications such as medical diagnosis, where evidence may be sparse, noisy, or conflicting. For example, DST can integrate results from various diagnostic tests and clinical symptoms to produce a unified and robust estimate of a patient's condition—even when some inputs are uncertain.

At the core of DST lies the Dempster's combination rule, which enables the fusion of multiple independent belief sources. Suppose we have two sets of mass functions, $\mathcal{S}^1 = \{\mathcal{B}_1^1, \mathcal{B}_2^1, \ldots, \mathcal{B}_K^1, \mathcal{U}^1\}$ and $\mathcal{S}^2 = \{\mathcal{B}_1^2, \mathcal{B}_2^2, \ldots, \mathcal{B}_K^2, \mathcal{U}^2\}$, each derived from distinct views. The fused mass $\mathcal{S} = \{\mathcal{S}_1, \mathcal{S}_2, \ldots, \mathcal{S}_K, \mathcal{U}\} = \mathcal{S}^1 \oplus \mathcal{S}^2$ is computed as:

$$\mathcal{B}_k = \frac{1}{1-C}(\mathcal{B}_k^1 \mathcal{B}_k^2 + \mathcal{B}_k^1 \mathcal{U}^2 + \mathcal{B}_k^2 \mathcal{U}^1), \mathcal{U} = \frac{1}{1-C}\mathcal{U}^1 \mathcal{U}^2, \tag{6}$$

where $\mathcal{B}_k^1$ and $\mathcal{B}_k^2$ are the belief masses assigned to class $k$ from view 1 and view 2, respectively. $\mathcal{U}^1$ and $\mathcal{U}^2$ are the uncertainty mass from each view, $C = \sum_{i \neq j} \mathcal{B}_i^1 \mathcal{B}_j^2$ is the conflict coefficient, quantifying the degree of disagreement between the two views.

The normalization factor $\frac{1}{1-C}$ ensures that the total mass of belief and uncertainty sums to 1, preserving probabilistic consistency. When the conflict $C$ is low, the combination rule amplifies agreement and effectively reduces uncertainty. In contrast, a high conflict score reflects strong disagreement, leading to less confident fused results.

Once belief and uncertainty mass have been computed for all available views, we iteratively apply the Dempster combination rule to aggregate them into a unified representation. This fused result captures the consensus across multiple perspectives, integrating their complementary strengths while mitigating the weaknesses or uncertainties of any single view.

In summary, DST provides a theoretically grounded and computationally practical mechanism for multi-view decision fusion, enabling our model to make more informed and robust predictions in the presence of uncertainty and incomplete information.

*3.3. Dynamic Multi-View Adaptive Decision Framework*

In this subsection, we propose a dynamic decision-making framework designed to optimize diagnostic performance and resource allocation through real-time evaluation of predictive uncertainty. As illustrated in Figure 3, the proposed architecture progressively integrates multi-view data in a cost-effective manner, ensuring diagnostic reliability while reducing the economic burden associated with acquiring extensive medical data.

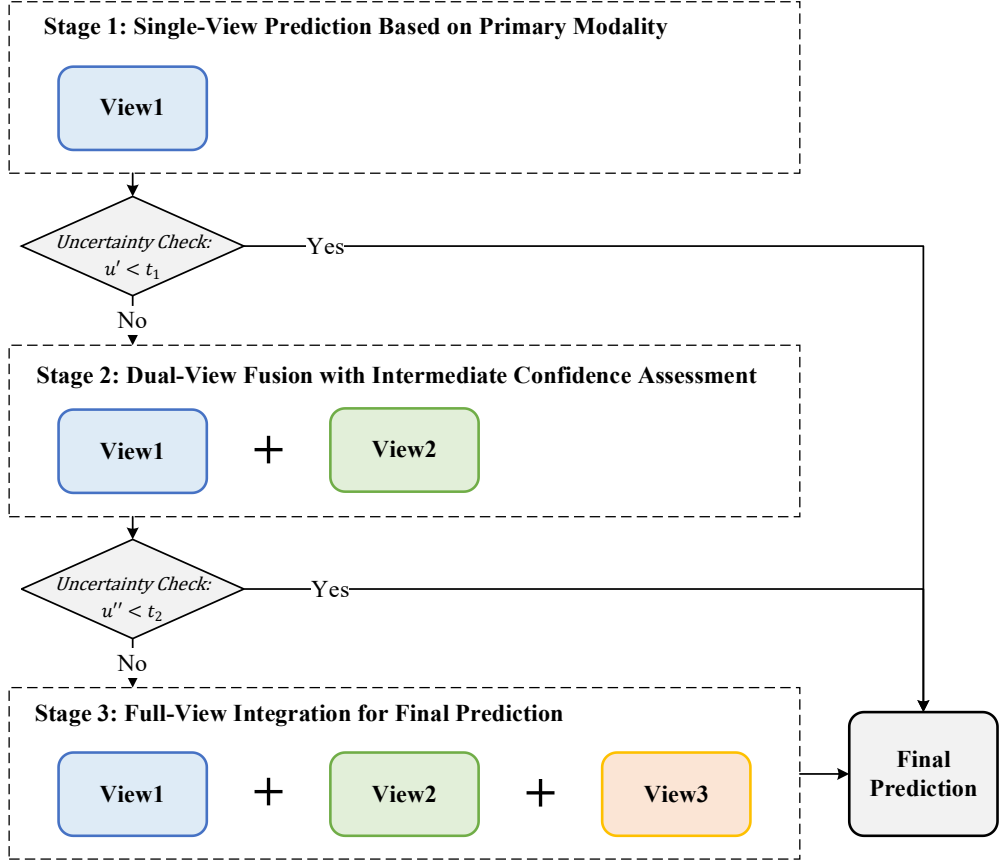

**Figure 3.** An uncertainty-aware multi-view dynamic decision framework for prediction.

**Dynamic Decision Mechanism:** The core of the framework lies in its multi-stage decision mechanism driven by uncertainty estimation. The model continuously monitors its prediction confidence and dynamically adjusts the input sources based on predefined uncertainty thresholds. When the uncertainty of a model's output exceeds a certain threshold, the system identifies the need to incorporate additional views to improve robustness. This progressive, uncertainty-aware data fusion continues until the predictive uncertainty is reduced to an acceptable level or until all available modalities have been exhausted.

**Three-Stage Workflow:** The framework operates through three sequential stages. Initially, a primary predictive model is constructed using single-view (e.g., baseline omics) data. A real-time uncertainty assessment module evaluates the model's output; if the uncertainty is below a calibrated threshold $t_1$, the prediction is accepted without further processing. If not, the system enters a second stage, where a secondary omics modality is integrated. At this point, the model applies Dempster–Shafer Theory (DST) to combine evidence from both views, yielding an updated confidence measure. A second threshold $t_2$ is used to assess this revised uncertainty. If it remains above $t_2$, the framework advances to the final stage. In this third phase, a comprehensive multi-view prediction is performed by integrating a third omics modality. This panoramic decision-making process leverages the full spectrum of available data, allowing for highly accurate and reliable diagnoses while avoiding the unnecessary collection of low-value data. By progressively integrating evidence in response to uncertainty, the model achieves both cost-efficiency and personalized decision support.

**Threshold Optimization Strategy**: A critical component of this adaptive framework is the selection of appropriate uncertainty thresholds, which dictate when additional views should be introduced. These thresholds are optimized through an iterative search process that explores different candidate values and evaluates model performance using metrics such as accuracy, F1-score, and ROC-AUC. The complete procedure is described in Algorithm 1.

---

**Algorithm 1**. Threshold Selection for Three Staged Multiomics Data Classification.

---

**Input:** Single-view uncertainty estimates: $u' = [u'_1, u'_2, ..., u'_n]$, Dual-view uncertainty estimates: $u'' = [u''_1, u''_2, ..., u''_n]$

**Output:** Optimal uncertainty thresholds: best_$t_1$ and best_$t_2$

1: Define threshold search ranges: range_$t_1$= linspace(min($u'$), max($u'$), 100), range_$t_2$= linspace(min($u''$), max($u''$), 100); Initialize best_$t_1$ = 0, best_$t_2$ = 0, and best_correct_count = 0
2: **for** each $t_1 \in$ range_$t_1$:
3:     **for** each $t_2 \in$ range_$t_2$:
4:         Initialize empty list: patients_result = []
5:         **for** each patient in dataset:
6:           **if** patient.$u' \leq t_1$:
7:             Assign classification based on single-view prediction: patients_result.append(single_view(patient))
8:           **else if** patient.$u'' \leq t_2$:
9:             Assign classification based on dual-view prediction: patients_result.append(dual_view(patient))
10:           **else**:
11:             Assign classification based on tri-view prediction: patients_result.append(tri_view(patient))
12:         **end for**
13:         Evaluate prediction accuracy: correct_count = count_correct(patients_result)
14:         **if** correct_count ≥ best_correct_count:
15:           Update best record: best_correct_count = correct_count, best_$t_1$= $t_1$, best_$t_2$ = $t_2$
16:         **end if**
17:     **end for**
18: **end for**

**Explanation of the Algorithm:** This algorithm aims to determine the optimal uncertainty thresholds $t_1$ and $t_2$ to guide the staged classification process. The inputs are the uncertainty estimates derived from single-view and dual-view models for each patient. To define the candidate thresholds, we uniformly sample 100 values between the minimum and maximum uncertainty levels observed in the single-view and dual-view settings, respectively. This creates two discrete search spaces for $t_1$ and $t_2$. An exhaustive grid search is then conducted across all possible threshold combinations to find the pair that maximizes classification performance.

For each threshold pair $(t_1, t_2)$, the algorithm evaluates every patient by first examining their single-view uncertainty. If this uncertainty is below $t_1$, the prediction is considered sufficiently reliable and directly accepted. If it exceeds $t_1$, the dual-view uncertainty is assessed. A dual-view uncertainty below $t_2$ leads to classification at this intermediate stage. Otherwise, the patient proceeds to the tri-view setting for further evaluation. After classifying all patients under the current threshold configuration, the number of correctly predicted cases is computed. If this value exceeds or matches the best result observed so far, the current thresholds are stored as the new optimum. The algorithm continues this evaluation until all candidate threshold pairs have been assessed. In the end, the threshold pair $(t_1, t_2)$ that delivers the highest overall classification accuracy is selected as the final output.

*3.4. Loss function*

During model training, the total loss function is composed of two primary components: the cross-entropy loss and a Kullback–Leibler (KL) divergence term. The cross-entropy loss encourages the model to assign higher evidence to the correct labels, while the KL divergence acts as a regularizer, suppressing the evidence associated with incorrect predictions and thereby promoting calibrated uncertainty estimates.

The standard cross-entropy loss, denoted as the objective for classifier $c$, is defined as:

$$\mathcal{L}_{ce} = -\sum_{j=1}^{K} g_{ij} \log(p_{ij}), \tag{7}$$

where $p_{ij}$ is the predicted probability that sample $i$ belongs to class $j$, and $g_{ij}$ is the one-hot encoded ground truth label. However, this conventional formulation does not account for predictive uncertainty, as it directly compares deterministic predictions with the target distribution.

To address this limitation, our model adopts an evidential learning framework in which each prediction is modeled as a Dirichlet distribution $\mathcal{D}(p_i \mid d_i)$, parameterized by evidence $e_i$. Accordingly, we redefine the cross-entropy loss as its expectation under the Dirichlet distribution:

$$\mathcal{L}_{ace}(d_i) = \int \left[\sum_{j=1}^{k} -g_{ij} \log(p_{ij})\right] \frac{1}{B(d_i)} \prod_{j=1}^{K} p_{ij}^{d_{ij}-1} \, dp_i = \sum_{j=1}^{K} g_{ij} \left(\psi(S_i) - \psi(d_{ij})\right), \quad (8)$$

where $T_i = \sum_{j=1}^{K} d_{ij}$, $\psi(\cdot)$ denotes the Digamma function, $B(\cdot)$ is the multivariate Beta function, and $S_i$ the Dirichlet strength. This formulation penalizes overconfident predictions on uncertain inputs, since the gap $\psi(S_i) - \psi(d_{ij})$ widens when the evidence is insufficient, enforcing better uncertainty calibration.

The KL divergence term regularizes the predictive distribution by measuring its distance from a non-informative uniform Dirichlet prior $\mathcal{D}(p_i|\mathbf{1})$, and is given by:

$$\text{KL}[\mathcal{D}(p_i|\tilde{d}_i) \parallel \mathcal{D}(p_i|\mathbf{1})] = \log\left(\frac{\Gamma(\sum_{k=1}^{K} \tilde{d}_{ik})}{\Gamma(K)\prod_{k=1}^{K}\Gamma(\tilde{d}_{ik})}\right) + \sum_{k=1}^{K}(\tilde{d}_{ik}-1)[\psi(\tilde{d}_{ik}) - \psi(\sum_{j=1}^{K}\tilde{d}_{ij})], \quad (9)$$

where $\tilde{d}_i = y_i + (1-y_i) \circ d_i$ is the adjusted Dirichlet parameter (Cover, 1999) ensuring that the true class evidence is preserved and not penalized. Additionally, $\Gamma(\cdot)$ denotes the Gamma function.

Given the Dirichlet parameter $d_i$ for the $i$-th sample, the full loss function is defined as:

$$\mathcal{L}(d_i) = \mathcal{L}_{ace}(d_i) + \eta_t \text{KL}[\mathcal{D}(p_i|\tilde{d}_i) \parallel \mathcal{D}(p_i|\mathbf{1})], \quad (10)$$

where $\eta_t > 0$ is a dynamic balancing coefficient that gradually increases from 0 to 1 during training. This scheduling prevents the KL divergence term from dominating the early optimization process, allowing the model to first focus on learning discriminative evidence.

For each sample $i$ with $V$ available views, the final training objective combines both single-view and multi-view losses, and is expressed as:

$$\mathcal{L}_{\text{overall}} = \sum_{i=1}^{N}\left[\mathcal{L}(d_i) + \sum_{v=1}^{V}\mathcal{L}(d_v^i)\right], \quad (11)$$

where $d_v^i$ denotes the Dirichlet parameter derived from the $v$-th view of sample $i$. This comprehensive formulation ensures both view-specific and fused predictions are jointly optimized under a unified evidential learning framework.

## 4. Experiments

### 4.1. Datasets

To evaluate the effectiveness of our model, we conducted experiments on four real-world multi-omics datasets. The results demonstrate the model's superior performance in addressing complex multi-omics classification tasks. The datasets used in this study are as follows:

- **ROSMAP** (A. Bennett, A. Schneider, Arvanitakis, & S. Wilson, 2012): This dataset comprises 351 clinical samples, including individuals diagnosed with Alzheimer's disease (AD) and healthy controls. It is widely adopted in studies focusing on early-stage dementia detection and neurodegenerative disease research.
- **LGG** (A Bennett, et al., 2012): Consisting of 510 brain glioma samples, this dataset includes 246 low-grade and 264 intermediate-grade tumors. It supports tumor malignancy classification and the study of glioma progression.
- **BRCA** (Cancer Genome Atlas Research Network, 2013): Comprising 876 breast cancer cases, this dataset categorizes samples into five molecular subtypes based on gene expression profiles. It is primarily used for molecular subtyping and personalized diagnostics.
- **KIPAN** (Cancer Genome Atlas Research Network, 2013): This dataset includes 658 samples of kidney tumors across three major renal carcinoma subtypes, facilitating differential diagnosis and cross-type comparative analysis.

To comprehensively capture the biological characteristics inherent in these datasets, we utilized three types of omics data, each representing distinct molecular layers:

- **mRNA** [28]: Reflects transcriptional activity and gene expression patterns. It provides insights into dysregulated genes and molecular mechanisms underlying disease phenotypes—for instance, aberrant expression in neurodegenerative disorders like Alzheimer's disease.
- **DNA methylation (DNAmethy)** [29]: Captures methylation status at CpG sites, revealing epigenetic regulation shaped by environmental and aging factors. It plays a key role in modulating gene expression and identifying silenced or activated regions in disease contexts.
- **miRNA** [30]: Represents post-transcriptional regulation through small non-coding RNA molecules. miRNAs are critical for gene silencing and have been implicated in controlling pathways associated with neurodegeneration, cancer progression, and cellular homeostasis.

By integrating these complementary omics modalities, our model is designed to exploit the hierarchical and cross-dimensional features that characterize complex diseases, thereby improving both diagnostic precision and biological interpretability.

## 4.2. Model implementation and omics selection

We constructed single-view, dual-view, and triple-view classification models using different combinations of omics data to assess the model's performance in multi-omics classification tasks.

- **Single-view models** were built using individual omics modalities, including mRNA expression, DNA methylation, and miRNA expression.
- **Dual-view models** integrated two omics types, such as mRNA + DNAmethy, mRNA + miRNA, and DNAmethy + miRNA.
- **Triple-view models** utilized all three omics modalities simultaneously—mRNA, DNAmethy, and miRNA—to form a comprehensive feature representation.

The detailed performance metrics of the proposed model under different omics combinations are presented in Table 1 (ROSMAP and LGG) and Table 2 (BRCA and KIPAN). These results enable a systematic comparison of classification performance across various omics views. For binary classification tasks, we adopted accuracy, F1-score, and area under the receiver operating characteristic curve (AUC) as evaluation metrics. For multi-class classification, we used accuracy, weighted F1-score (Weighted-F1), and macro-average F1-score (Macro-F1).

**Table 1.** Classification performance and average uncertainty using the proposed method in ROSMAP and LGG.

| Dataset | Omics View | Accuracy | F1-score | AUC | Uncertainty |
|---|---|---|---|---|---|
| ROSMAP (A. Bennett, et al., 2012) | mRNA | 0.820 | 0.831 | 0.819 | 0.518 |
| | DNAmethy | 0.622 | 0.729 | 0.608 | 0.427 |
| | miRNA | 0.707 | 0.752 | 0.701 | 0.491 |
| | mRNA+DNAmethy | 0.839 | 0.849 | 0.838 | 0.426 |
| | mRNA+miRNA | 0.820 | 0.834 | 0.818 | 0.441 |
| | DNAmethy+miRNA | 0.745 | 0.787 | 0.738 | 0.429 |
| | mRNA+DNAmethy+miRNA | 0.858 | 0.873 | 0.855 | 0.302 |
| LGG (A Bennett, et al., 2012) | mRNA | 0.830 | 0.816 | 0.833 | 0.323 |
| | DNAmethy | 0.705 | 0.621 | 0.713 | 0.484 |
| | miRNA | 0.738 | 0.726 | 0.740 | 0.398 |
| | mRNA+DNAmethy | 0.836 | 0.827 | 0.839 | 0.192 |
| | mRNA+miRNA | 0.843 | 0.837 | 0.845 | 0.258 |
| | DNAmethy+miRNA | 0.758 | 0.729 | 0.762 | 0.428 |
| | mRNA+DNAmethy+miRNA | 0.856 | 0.849 | 0.858 | 0.119 |

**Table 2.** Classification performance and average uncertainty using the proposed method in BRCA and KIPAN.

| Dataset | Omics View | Accuracy | F1-score | Micro-F1 | Uncertainty |
|---|---|---|---|---|---|
| BRCA (Cancer Genome Atlas Research Network, 2013) | mRNA | 0.866 | 0.870 | 0.837 | 0.496 |
| | DNAmethy | 0.718 | 0.723 | 0.678 | 0.599 |
| | miRNA | 0.752 | 0.750 | 0.689 | 0.610 |
| | mRNA+DNAmethy | 0.851 | 0.853 | 0.823 | 0.414 |
| | mRNA+miRNA | 0.866 | 0.866 | 0.833 | 0.401 |
| | DNAmethy+miRNA | 0.794 | 0.788 | 0.744 | 0.541 |
| | mRNA+DNAmethy+miRNA | 0.855 | 0.859 | 0.821 | 0.357 |
| KIPAN (Cancer Genome Atlas Research Network, 2013) | mRNA | 0.924 | 0.921 | 0.871 | 0.147 |
| | DNAmethy | 0.994 | 0.994 | 0.996 | 0.199 |
| | miRNA | 0.969 | 0.969 | 0.951 | 0.185 |
| | mRNA+DNAmethy | 1.000 | 1.000 | 1.000 | 0.058 |
| | mRNA+miRNA | 0.934 | 0.93.1 | 0.889 | 0.084 |
| | DNAmethy+miRNA | 0.974 | 0.97.4 | 0.974 | 0.175 |
| | mRNA+DNAmethy+miRNA | 1.000 | 1.000 | 1.000 | 0.034 |

We first evaluated single-view models, where the classification performance varied depending on the biological informativeness of each omics type. On the ROSMAP, LGG, and BRCA datasets, mRNA-based models consistently achieved superior performance compared to those based on DNA methylation and miRNA, with the highest classification accuracies of 0.820, 0.830, and 0.866, respectively. These models also outperformed others in terms of F1-score and AUC (see mRNA rows in Tables 1 and 2). In contrast, the KIPAN dataset showed a different pattern, where DNA methylation-based models achieved the highest accuracy of 0.994, outperforming the mRNA-based model by more than 6 percentage points.

Next, we developed dual-view models by combining two omics modalities. These combinations often led to notable improvements in classification accuracy. For example, the mRNA + miRNA model achieved 0.843 accuracy on LGG and 0.866 on BRCA (see mRNA+miRNA in Tables 1 and 2). However, the optimal dual-omics combination differed across datasets. In ROSMAP and KIPAN, combining mRNA with DNA methylation produced the best results, with 0.839 and 1.000 accuracy, respectively (see mRNA+DNAmethy in Tables 1 and 2). The perfect performance in KIPAN underscores the complementary nature of mRNA and methylation signals in certain cancer types.

Finally, we constructed tri-view models that integrate all three omics layers. These models consistently achieved strong results, demonstrating their ability to capture complex, multilayer biological patterns. The tri-view models (see mRNA+DNAmethy+miRNA in Tables 1 and 2) attained accuracies of 0.858 (ROSMAP), 0.856 (LGG), 0.855 (BRCA), and 1.000 (KIPAN). By synthesizing information across transcriptional, epigenetic, and post-transcriptional layers, the tri-view models offer a more holistic representation of disease mechanisms and significantly improve classification accuracy and robustness.

### 4.3. Model performance and comparison

We conducted extensive experiments on the four widely used datasets described above, analyzing each sample using mRNA expression, DNA methylation, and miRNA expression data. The proposed method was compared against 14 representative classification algorithms, encompassing both traditional machine learning approaches and state-of-the-art multi-omics integration methods. The comparative results are summarized in Table 3.

The baseline methods include:

- K-Nearest Neighbors (KNN) (Fix, 1985), Support Vector Machine (SVM) (Vapnik, 1995), L1-Regularized Linear Regression (LR) (Tibshirani, 1996), Random Forest (RF) (Ho, 1995), and Fully Connected Neural Networks (NN) (McCulloch & Pitts, 1943): These classical algorithms utilize diverse paradigms such as distance-based decision making, hyperplane separation, sparse modeling, ensemble learning, and deep representation learning.
- Adaptive Group-Regularized Ridge Regression (GRridge) (Van De Wiel, Lien, Verlaat, van Wieringen, & Wilting, 2016): A ridge regression extension tailored for high-dimensional data, employing adaptive group-specific penalties to enhance feature selection and classification performance.
- Block Partial Least Squares Discriminant Analysis (BPLSDA) and Block Sparse Partial Least Squares Discriminant Analysis (BSPLSDA) (Singh, et al., 2019): BPLSDA adapts Sparse Generalized Canonical Correlation Analysis (Sparse GCCA) for classification tasks, while BSPLSDA introduces additional sparsity constraints to refine feature selection and improve model interpretability.
- Multi-Omics Graph Convolutional Networks (MOGONET) (Wang, et al., 2021): A multi-omics integration framework that leverages graph convolutional networks (GCNs) to capture omics-specific structural information and a view correlation discovery network (VCDN) to model cross-omics relationships for classification.
- Trusted Multi-View Classification (TMC) (Han, Zhang, et al., 2022): A dynamic multi-view approach that evaluates the reliability of each modality at the individual sample level, integrating them in a confidence-aware manner for robust classification.
- Concatenation-Based Feature Fusion (CF) (Hong, et al., 2020): A late-fusion strategy that integrates multimodal representations through compact feature fusion techniques, ensuring effective cross-modality information integration.
- Gated Multimodal Units (GMU) (Arevalo, Solorio, Montes-y-Gómez, & González, 2017): A gating mechanism that selectively controls the flow of modality-specific information, enhancing discriminative feature fusion.
- Multi-Modality Dynamic Fusion (MMDynamic) (Han, Yang, Huang, Zhang, & Yao, 2022): A dynamic fusion approach that incorporates both feature-level and modality-specific gating mechanisms to adaptively integrate multi-source data while preserving inter-modality dependencies.

- Multi-Level Confidence Learning Network (MLCLNet) (Zheng, Tang, Wan, Hu, & Zhang, 2023): A framework that employs confidence-based feature selection to suppress redundant information and utilizes GCNs to model the underlying structure of multimodal data, enabling effective multi-view information fusion.

**Table 3.** Comparison with state-of-the-art methods on multi-omics classification. Bold text indicates the best performance.

| Model | ROSMAP (A. Bennett, et al., 2012) | | | LGG (A Bennett, et al., 2012) | | |
|---|---|---|---|---|---|---|
| | Accuracy | F1-score | AUC | Accuracy | F1-score | AUC |
| KNN (Fix, 1985) | 0.657±0.036 | 0.671±0.045 | 0.709±0.045 | 0.729±0.034 | 0.738±0.038 | 0.799±0.038 |
| SVM (Vapnik, 1995) | 0.770±0.024 | 0.778±0.026 | 0.770±0.026 | 0.754±0.046 | 0.757±0.046 | 0.754±0.046 |
| LR (Tibshirani, 1996) | 0.694±0.037 | 0.730±0.035 | 0.770±0.035 | 0.761±0.018 | 0.767±0.027 | 0.823±0.027 |
| RF (Ho, 1995) | 0.726±0.029 | 0.734±0.019 | 0.811±0.019 | 0.748±0.012 | 0.742±0.010 | 0.823±0.010 |
| NN (McCulloch & Pitts, 1943) | 0.755±0.021 | 0.764±0.025 | 0.827±0.025 | 0.737±0.023 | 0.748±0.037 | 0.810±0.037 |
| GRridge (Van De Wiel, et al., 2016) | 0.760±0.034 | 0.769±0.023 | 0.841±0.023 | 0.746±0.038 | 0.756±0.044 | 0.826±0.044 |
| BPLSDA (Singh, et al., 2019) | 0.742±0.024 | 0.755±0.025 | 0.830±0.025 | 0.759±0.025 | 0.738±0.023 | 0.825±0.023 |
| BSPLSDA (Singh, et al., 2019) | 0.753±0.033 | 0.764±0.021 | 0.838±0.021 | 0.685±0.027 | 0.662±0.026 | 0.730±0.026 |
| MOGONET (Wang, et al., 2021) | 0.815±0.023 | 0.821±0.012 | 0.874±0.012 | 0.816±0.016 | 0.814±0.027 | 0.840±0.027 |
| TMC (Han, Zhang, et al., 2022) | 0.825±0.009 | 0.823±0.006 | 0.885±0.006 | 0.819±0.008 | 0.815±0.004 | 0.871±0.004 |
| CF (Hong, et al., 2020) | 0.784±0.011 | 0.788±0.005 | 0.880±0.005 | 0.811±0.012 | 0.822±0.004 | 0.881±0.004 |
| GMU (Arevalo, et al., 2017) | 0.776±0.025 | 0.784±0.016 | 0.869±0.016 | 0.803±0.015 | 0.808±0.012 | 0.886±0.012 |
| MMDynamic (Han, Yang, et al., 2022) | 0.842±0.013 | 0.846±0.007 | **0.912±0.007** | 0.833±0.010 | 0.837±0.004 | **0.885±0.004** |
| MLCLNet (Zheng, et al., 2023) | 0.844±0.015 | 0.852±0.015 | 0.893±0.011 | 0.835±0.014 | 0.840±0.013 | 0.886±0.012 |
| **Proposed** | **0.858±0.012** | **0.873±0.011** | 0.855±0.012 | **0.856±0.010** | **0.849±0.012** | 0.858±0.017 |

**Evaluation Metrics.** We evaluated all models on binary classification tasks (ROSMAP and LGG datasets), using three performance metrics: accuracy, F1-score, and area under the ROC curve (AUC). These metrics jointly reflect the models' predictive power, robustness to class imbalance, and generalization ability.

**Results and Analysis.** Table 3 presents a comprehensive comparison of classification performance across multiple state-of-the-art models on the ROSMAP and LGG datasets. Among traditional machine learning methods, KNN consistently showed the weakest performance, particularly on the ROSMAP dataset (Accuracy: 0.657, AUC: 0.709), confirming its limitations in high-dimensional, heterogeneous multi-omics scenarios. SVM and LR performed moderately, achieving Accuracy values in the 0.694–0.770 range. Although RF and NN showed improved performance, with ROSMAP AUCs of 0.811 and o.827 respectively, their effectiveness remained constrained by their inability to fully capture the interdependencies across modalities.

Advanced multi-omics integration methods demonstrated clear advantages. MOGONET, which leverages modality-specific graph convolutional networks (GCNs) and a view correlation discovery network, achieved notable gains (ROSMAP AUC: 0.874). TMC and GMU, both of which emphasize dynamic or gated fusion mechanisms, achieved consistent performance, with TMC obtaining an AUC of 0.885 and GMU reaching 0.869 on ROSMAP. On the LGG dataset, both models maintained high AUCs above 0.871, showcasing reasonable generalization.

The CF method, while simple, showed surprisingly strong results (ROSMAP AUC: 0.880, LGG AUC: 0.881), indicating that even linear fusion methods can be effective when omics signals are relatively separable. However, their performance may deteriorate under more complex inter-modality conditions due to lack of adaptive weighting or structural modeling.

MMDynamic and MLCLNet—two of the strongest recent baselines—exhibited high performance across both datasets. MMDynamic achieved the highest AUC on ROSMAP (0.912), likely due to its multi-level attention mechanism and dynamic modality selection. MLCLNet performed comparably, with an F1-score of 0.852 and an AUC of 0.893. However, both models demonstrated slight degradation on the LGG dataset (MMDynamic AUC: 0.885; MLCLNet AUC: 0.886), suggesting potential overfitting to dataset-specific modality characteristics.

In contrast, our proposed method achieved the highest Accuracy and F1-score on both datasets—0.858 Accuracy and 0.873 F1-score on ROSMAP, and 0.856 Accuracy and 0.849 F1-score on LGG—while maintaining high AUC values (0.855 and 0.858, respectively). Notably, our model exhibited the smallest performance gap between datasets, which underscores its strong generalizability and robustness to cross-cohort variability. This consistency is particularly valuable in biomedical applications, where patient heterogeneity and modality noise are prevalent.

Moreover, unlike many existing methods that either require complex architecture tuning (e.g.,

GMU, MLCLNet) or involve multi-stage training (e.g., MOGONET), our model integrates feature extraction, modality-aware attention, and multi-view fusion in a unified, end-to-end framework. This not only simplifies deployment but also reduces potential error propagation between stages.

In summary, the results clearly demonstrate that our method is state-of-the-art in both accuracy and stability, outperforming both classical and advanced integration approaches. Its superiority stems from its ability to capture complementary information across omics modalities while maintaining robust performance across diverse clinical datasets.

*4.4. Optimal Threshold Selection for Improved Cost-Effectiveness*

We further analyzed the uncertainty distribution of correct and incorrect predictions produced by our model, as illustrated in Figure 4. It can be observed that correct predictions are typically associated with significantly lower uncertainty, indicating the model's high confidence when making accurate judgments. This suggests that the model effectively captures critical data features when predictions are reliable. Such characteristics are of great practical value, particularly in clinical scenarios where interpretability and risk assessment are crucial. For samples exhibiting high uncertainty, it becomes necessary to adopt a more cautious decision-making process.

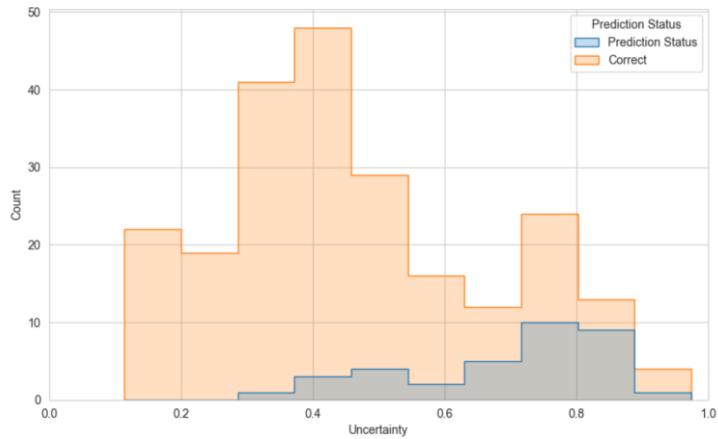

**Figure 4.** Histogram of uncertainty distribution for correct and incorrect predictions.

To address this, we introduce a progressive strategy guided by two empirically determined uncertainty thresholds: the first threshold $t_1$ is used to identify cases that can be reliably classified using only a single omics modality, while the second threshold $t_2$ determines when predictions require integration of all available omics views to ensure reliability. These thresholds are selected based on optimization of classification performance and are dataset-specific. The corresponding values of $t_1$ and $t_2$ for each of the four benchmark datasets are presented in Table 4.

Table 4. Optimal uncertainty thresholds $t_1$ and $t_2$ for each dataset.

| Dataset | $t_1$ | $t_2$ |
|---|---|---|
| ROSMAP (A. Bennett, et al., 2012) | 0.593 | 0.465 |
| LGG (A Bennett, et al., 2012) | 0.319 | 0.239 |
| BRCA (Cancer Genome Atlas Research Network, 2013) | 0.514 | 0.856 |
| KIPAN (Cancer Genome Atlas Research Network, 2013) | 0.321 | 0.330 |

These optimized thresholds allow the model to achieve high-confidence predictions in many cases using only one or two omics modalities, thereby reducing the need for full multi-omics integration and offering improved cost-effectiveness. Based on these thresholds, the distribution of samples across the three decision stages is reported in Table 5, which summarizes the proportion of cases handled at each stage and the omics views involved.

Table 5. Sample distribution across decision stages for multi-omics classification.

| Dataset | Stage 1 (Single View) | Stage 2 (Dual View) | Stage 3 (Triple View) |
|---|---|---|---|
| ROSMAP (A. Bennett, et al., 2012) | 68.87%(mRNA) | 2.83%(mRNA+DNAmethy) | 28.30%(mRNA+miRNA+DNAmethy) |
| LGG (A Bennett, et al., 2012) | 47.06%(mRNA) | 3.27%(mRNA+miRNA) | 49.67%(mRNA+miRNA+DNAmethy) |
| BRCA (Cancer Genome Atlas Research Network, 2013) | 60.08%(mRNA) | 39.92%(mRNA+miRNA) | 0.00% |
| KIPAN (Cancer Genome Atlas Research Network, 2013) | 92.42% (DNAmethy) | 7.58%(mRNA+DNAmethy) | 0.00% |

From Table 5, it can be observed that, except for the LGG dataset, a single omics modality was sufficient to generate reliable predictions for the majority of patients across the other three datasets. Even in the relatively complex LGG dataset, over 47.06% of the cases were confidently classified using only one type of omics data. Notably, our model demonstrated exceptional performance on the KIPAN dataset, where more than 92% of predictions were made using solely DNA methylation data. Furthermore, both the KIPAN and BRCA datasets required no full three-view integration, as evidenced by the absence of samples in Stage 3 for these datasets, indicating that all reliable predictions were completed at earlier decision stages.

These findings underscore the practicality of our stage-wise framework, where a vast majority of predictions can be accurately made using only a single omics modality. This not only streamlines the diagnostic process but also significantly reduces financial and computational costs. This observation is further supported by the fact that Stage 3 values for LGG and KIPAN are 0.00% in Table 5, indicating that all predictions for these datasets were completed before reaching the final integration stage. Meanwhile, for a smaller subset of cases characterized by higher predictive uncertainty, additional omics views can be incorporated to support more in-depth analysis and improve confidence in the final diagnosis.

Within our framework, prediction begins with a single omics modality. Only when the model encounters high uncertainty does it selectively introduce additional omics views, enabling a progressive refinement of the prediction. This stage-wise strategy ensures that computational and economic resources are allocated primarily to uncertain cases, thereby improving overall model efficiency and reliability.

Our results reveal a clever trade-off between diagnostic accuracy and cost-effectiveness. The model leverages single-view data to rapidly and reliably classify the majority of patients, minimizing both resource usage and patient burden. When needed, a second omics modality is added to bolster predictions for uncertain cases. Ultimately, only a small fraction of highly uncertain samples proceed to Stage 3, requiring full multi-omics integration. In this way, the model achieves a balanced and adaptive use of data, ensuring both accuracy and affordability in clinical decision-making. The effectiveness of this framework is further confirmed by the uncertainty distributions observed at each stage, which are summarized in Figure 5.

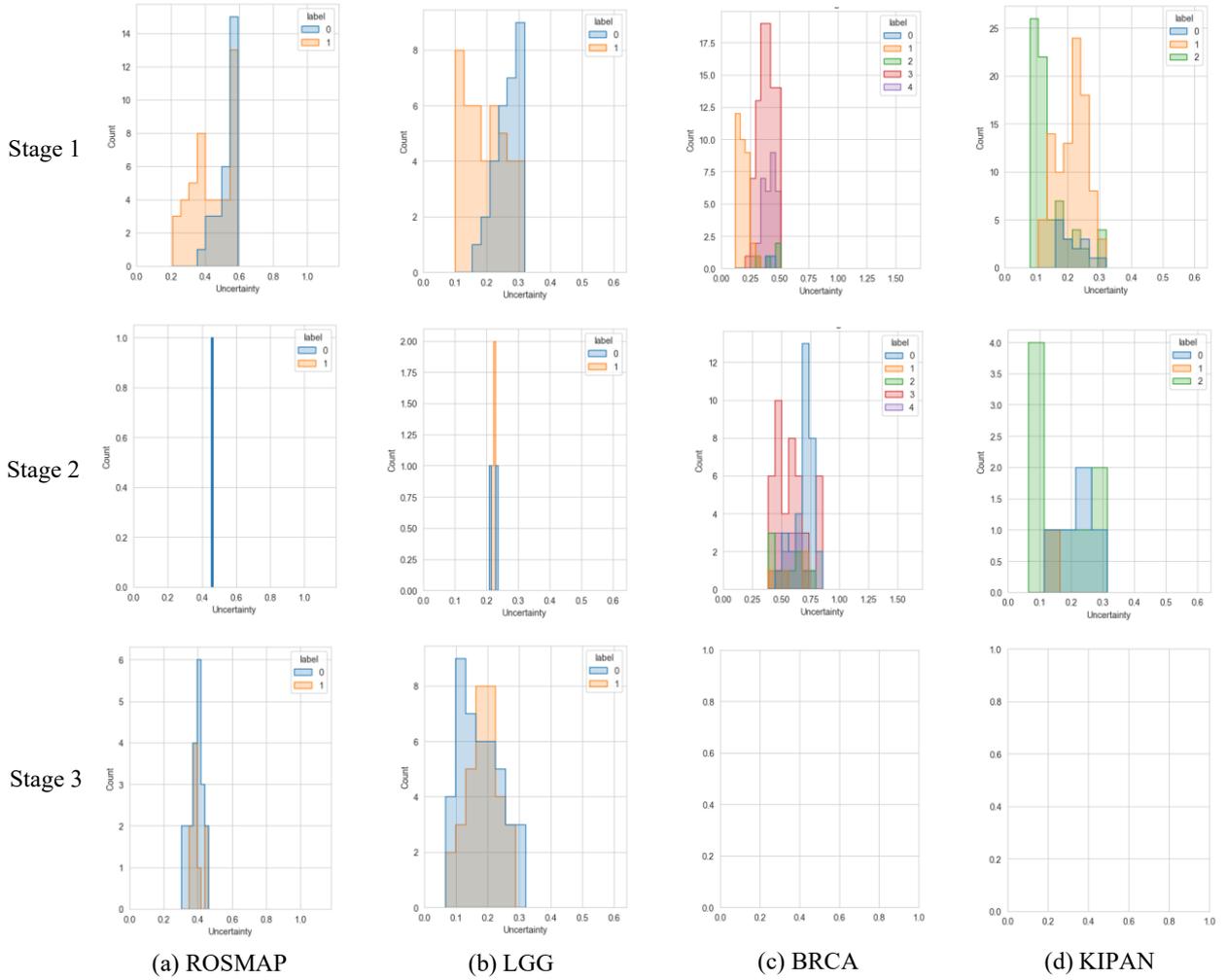

**Figure 5.** Distribution of classification uncertainty across three decision stages for four datasets.

As shown in Figure 5, across all datasets, the majority of samples lie within the low-uncertainty region, indicating the model's high confidence in its predictions and supporting the effectiveness of using single-view data in most cases. Although integrating more omics views can improve prediction in certain situations, it may also introduce additional noise or redundancy, occasionally increasing uncertainty. Our dynamic decision-making framework effectively balances diagnostic accuracy and cost-efficiency, ensuring optimal use of multi-omics information. Notably, the absence of Stage 3 samples in both the BRCA and KIPAN datasets further illustrates the efficiency of our approach, as reliable predictions were achieved using only two omics types, highlighting the model's ability to maintain diagnostic performance while avoiding unnecessary medical or computational expenses.

## 5. Conclusion

This study presents an uncertainty-driven dynamic decision-making framework for the progressive integration of high-throughput omics data. Unlike conventional approaches that require the collection of all omics data before making a decision, our method quantifies the classification uncertainty of single-view models at the evidence level and employs an adaptive thresholding mechanism based on Dempster-Shafer theory to dynamically determine when additional omics modalities are necessary. Only when the model exhibits insufficient confidence are supplementary views incorporated. Experiments on four publicly available datasets show that, in three of them, over half of the samples achieve reliable predictions without requiring full multi-omics input—effectively reducing patient costs while preserving diagnostic accuracy. For more complex cases, the selective integration of additional omics data further enhances prediction robustness. These results highlight the clinical applicability of our framework and propose a novel, cost-effective paradigm for the efficient utilization of multi-omics information in precision medicine.

**Acknowledgment**

Nan Mu was supported by the Natural Science Foundation of Sichuan Province (grant number 2025ZNSFSC1477) and the National Natural Science Foundation of China (grant number 62006165).

# References

A Bennett, D., A Schneider, J., S Buchman, A., L Barnes, L., A Boyle, P., & S Wilson, R. (2012). Overview and findings from the rush Memory and Aging Project. *Current Alzheimer Research, 9*, 646-663.

A. Bennett, D., A. Schneider, J., Arvanitakis, Z., & S. Wilson, R. (2012). Overview and findings from the religious orders study. *Current Alzheimer Research, 9*, 628-645.

Arevalo, J., Solorio, T., Montes-y-Gómez, M., & González, F. A. (2017). Gated multimodal units for information fusion. *arXiv preprint arXiv:1702.01992*.

Cancer Genome Atlas Research Network, J. (2013). The cancer genome atlas pan-cancer analysis project. *Nat. Genet, 45*, 1113-1120.

Chen, M., Gao, J., & Xu, C. (2024). Revisiting Essential and Nonessential Settings of Evidential Deep Learning. *arXiv preprint arXiv:2410.00393*.

Cover, T. M. (1999). *Elements of information theory*: John Wiley & Sons.

Dar, M. A., Arafah, A., Bhat, K. A., Khan, A., Khan, M. S., Ali, A., Ahmad, S. M., Rashid, S. M., & Rehman, M. U. (2023). Multiomics technologies: role in disease biomarker discoveries and therapeutics. *Briefings in Functional Genomics, 22*, 76-96.

Dempster, A. P. (2008). Upper and lower probabilities induced by a multivalued mapping. In *Classic works of the Dempster-Shafer theory of belief functions* (pp. 57-72): Springer.

Deng, D., Chen, G., Yu, Y., Liu, F., & Heng, P.-A. (2023). Uncertainty estimation by fisher information-based evidential deep learning. In *International conference on machine learning* (pp. 7596-7616): PMLR.

Espinel-Ríos, S., López, J. M., & Avalos, J. L. (2025). Omics-driven hybrid dynamic modeling of bioprocesses with uncertainty estimation. *Biochemical Engineering Journal, 216*, 109637.

Fix, E. (1985). *Discriminatory analysis: nonparametric discrimination, consistency properties* (Vol. 1): USAF school of Aviation Medicine.

Flores, J. E., Claborne, D. M., Weller, Z. D., Webb-Robertson, B.-J. M., Waters, K. M., & Bramer, L. M. (2023). Missing data in multi-omics integration: Recent advances through artificial intelligence. *Frontiers in artificial intelligence, 6*, 1098308.

Fouché, A., & Zinovyev, A. (2023). Omics data integration in computational biology viewed through the prism of machine learning paradigms. *Frontiers in Bioinformatics, 3*, 1191961.

Gal, Y., & Ghahramani, Z. (2016). Dropout as a bayesian approximation: Representing model uncertainty in deep learning. In *international conference on machine learning* (pp. 1050-1059): PMLR.

Han, Z., Yang, F., Huang, J., Zhang, C., & Yao, J. (2022). Multimodal dynamics: Dynamical fusion for trustworthy multimodal classification. In *Proceedings of the IEEE/CVF conference on computer vision and pattern recognition* (pp. 20707-20717).

Han, Z., Zhang, C., Fu, H., & Zhou, J. T. (2022). Trusted multi-view classification with dynamic evidential fusion. *IEEE transactions on pattern analysis and machine intelligence, 45*, 2551-2566.

Hasan, M., Hossain, I., Rahman, A., & Nahavandi, S. (2023). Controlled dropout for uncertainty estimation. In *2023 IEEE International Conference on Systems, Man, and Cybernetics (SMC)* (pp. 973-980): IEEE.

Hasin, Y., Seldin, M., & Lusis, A. (2017). Multi-omics approaches to disease. *Genome biology, 18*, 1-15.

Ho, T. K. (1995). Random decision forests. In *Proceedings of 3rd international conference on document analysis and recognition* (Vol. 1, pp. 278-282): IEEE.

Hong, D., Gao, L., Yokoya, N., Yao, J., Chanussot, J., Du, Q., & Zhang, B. (2020). More diverse means better: Multimodal deep learning meets remote-sensing imagery classification. *IEEE Transactions on Geoscience and Remote Sensing, 59*, 4340-4354.

Jøsang, A. (2016). *Subjective logic* (Vol. 3): Springer.

Kendall, A., & Gal, Y. (2017). What uncertainties do we need in bayesian deep learning for computer vision? *Advances in neural information processing systems, 30*.

Ma, S., Zeng, A. G., Haibe-Kains, B., Goldenberg, A., Dick, J. E., & Wang, B. (2024). Integrate Any Omics: Towards genome-wide data integration for patient stratification. *arXiv preprint arXiv:2401.07937*.

MacKay, D. J. (1992). A practical Bayesian framework for backpropagation networks. *Neural computation, 4*, 448-472.
McCulloch, W. S., & Pitts, W. (1943). A logical calculus of the ideas immanent in nervous activity. *The bulletin of mathematical biophysics, 5*, 115-133.
Mobiny, A., Yuan, P., Moulik, S. K., Garg, N., Wu, C. C., & Van Nguyen, H. (2021). Dropconnect is effective in modeling uncertainty of bayesian deep networks. *Scientific reports, 11*, 5458.
Mullin, E. (2022). The era of fast, cheap genome sequencing is here. *Wired https://go.nature. com/3ofIjRH*.
Neal, R. M. (2012). *Bayesian learning for neural networks* (Vol. 118): Springer Science & Business Media.
Orsini, A., Diquigiovanni, C., & Bonora, E. (2023). Omics technologies improving breast cancer research and diagnostics. *International Journal of Molecular Sciences, 24*, 12690.
Picard, M., Scott-Boyer, M.-P., Bodein, A., Périn, O., & Droit, A. (2021). Integration strategies of multi-omics data for machine learning analysis. *Computational and Structural Biotechnology Journal, 19*, 3735-3746.
Sensoy, M., Kaplan, L., & Kandemir, M. (2018). Evidential deep learning to quantify classification uncertainty. *Advances in neural information processing systems, 31*.
Singh, A., Shannon, C. P., Gautier, B., Rohart, F., Vacher, M., Tebbutt, S. J., & Lê Cao, K.-A. (2019). DIABLO: an integrative approach for identifying key molecular drivers from multi-omics assays. *Bioinformatics, 35*, 3055-3062.
Song, J., Wang, C., Zhao, T., Zhang, Y., Xing, J., Zhao, X., Zhang, Y., & Zhang, Z. (2025). Multi-omics approaches for biomarker discovery and precision diagnosis of prediabetes. *Frontiers in Endocrinology, 16*, 1520436.
Tibshirani, R. (1996). Regression shrinkage and selection via the lasso. *Journal of the Royal Statistical Society Series B: Statistical Methodology, 58*, 267-288.
Van De Wiel, M. A., Lien, T. G., Verlaat, W., van Wieringen, W. N., & Wilting, S. M. (2016). Better prediction by use of co-data: adaptive group-regularized ridge regression. *Statistics in medicine, 35*, 368-381.
Vapnik, V. (1995). Support-vector networks. *Machine learning, 20*, 273-297.
Vitorino, R. (2024). Transforming clinical research: the power of high-throughput omics integration. *Proteomes, 12*, 25.
Wang, T., Shao, W., Huang, Z., Tang, H., Zhang, J., Ding, Z., & Huang, K. (2021). MOGONET integrates multi-omics data using graph convolutional networks allowing patient classification and biomarker identification. *Nature communications, 12*, 3445.
Wu, Y., & Xie, L. (2024). AI-driven multi-omics integration for multi-scale predictive modeling of causal genotype-environment-phenotype relationships. *arXiv preprint arXiv:2407.06405*.
Zhang, J. (2024). Emerging Trends in Multi-Omics Data Integration: Challenges and Future Directions. *Computational Molecular Biology, 14*.
Zhao, C., Liu, A., Zhang, X., Cao, X., Ding, Z., Sha, Q., Shen, H., Deng, H.-W., & Zhou, W. (2024). CLCLSA: Cross-omics linked embedding with contrastive learning and self attention for integration with incomplete multi-omics data. *Computers in biology and medicine, 170*, 108058.
Zheng, X., Tang, C., Wan, Z., Hu, C., & Zhang, W. (2023). Multi-level confidence learning for trustworthy multimodal classification. In *Proceedings of the AAAI conference on artificial intelligence* (Vol. 37, pp. 11381-11389).